\documentclass[11pt, a4paper]{article}

\usepackage[utf8]{inputenc}
\usepackage[T1]{fontenc}
\usepackage[english]{babel}
\usepackage{geometry}
\usepackage{graphicx}
\usepackage{amsmath}
\usepackage{amssymb}
\usepackage{caption}
\usepackage{subcaption}
\usepackage{hyperref}
\usepackage{times}
\usepackage{array}
\usepackage{tikz}
\usetikzlibrary{shapes.geometric, arrows, positioning, fit, calc}
\usepackage[toc, page]{appendix}
\usepackage{eso-pic}
\usepackage{transparent}

\tikzstyle{startstop} = [rectangle, rounded corners, minimum width=3cm, minimum height=1cm, text centered, draw=black, fill=blue!10]
\tikzstyle{process} = [rectangle, minimum width=3cm, minimum height=1cm, text centered, text width=3cm, draw=black, fill=orange!10]
\tikzstyle{model} = [cylinder, shape border rotate=90, aspect=0.25, minimum height=2.5cm, text centered, draw=black, fill=green!10]
\tikzstyle{data} = [trapezium, trapezium left angle=70, trapezium right angle=110, minimum width=2.5cm, minimum height=1cm, text centered, draw=black, fill=red!10]
\tikzstyle{arrow} = [thick,->,>=stealth]
\tikzstyle{line} = [thick,-]

\hypersetup{
    colorlinks=true, linkcolor=blue,
    filecolor=black, urlcolor=blue, citecolor=black,
}
\graphicspath{{images/}}

\title{\vspace{-2cm}\textbf{Proportion and Perspective Control for Flow-Based Image Generation}}
\author{
    Julien Boudier, Hugo Caselles-Dupré \\
    Obvious Research \\
    \texttt{research.obvious@gmail.com} \\
    Paris, France
}
\date{}

\begin{document}
\maketitle

\begin{figure}[h!]
    \centering
    \begin{subfigure}[b]{0.48\textwidth}
        \centering
        \includegraphics[width=\linewidth]{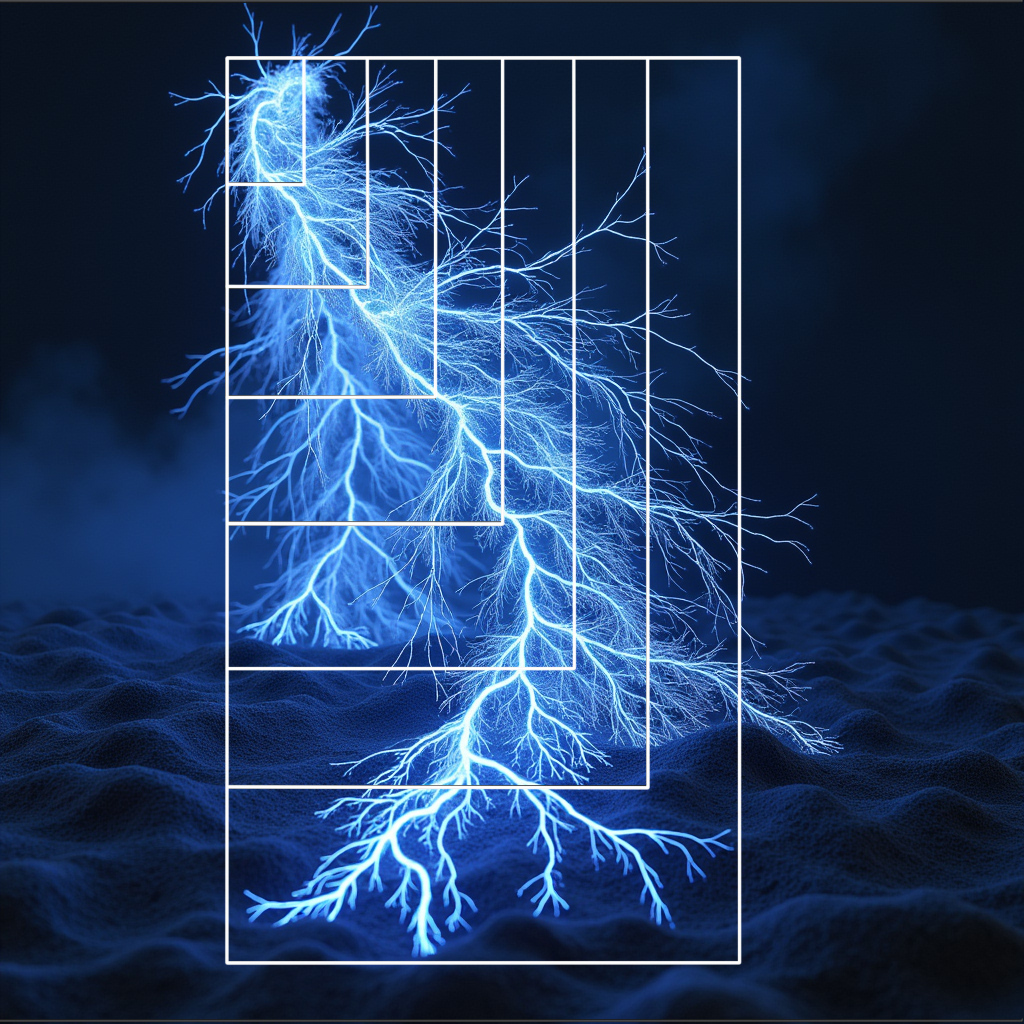}
        \caption{The Proportion ControlNet uses bounding boxes to guide object placement with proportions principles.}
        \label{fig:teaser_proportion}
    \end{subfigure}
    \hfill
    \begin{subfigure}[b]{0.48\textwidth}
        \centering
        \includegraphics[width=\linewidth]{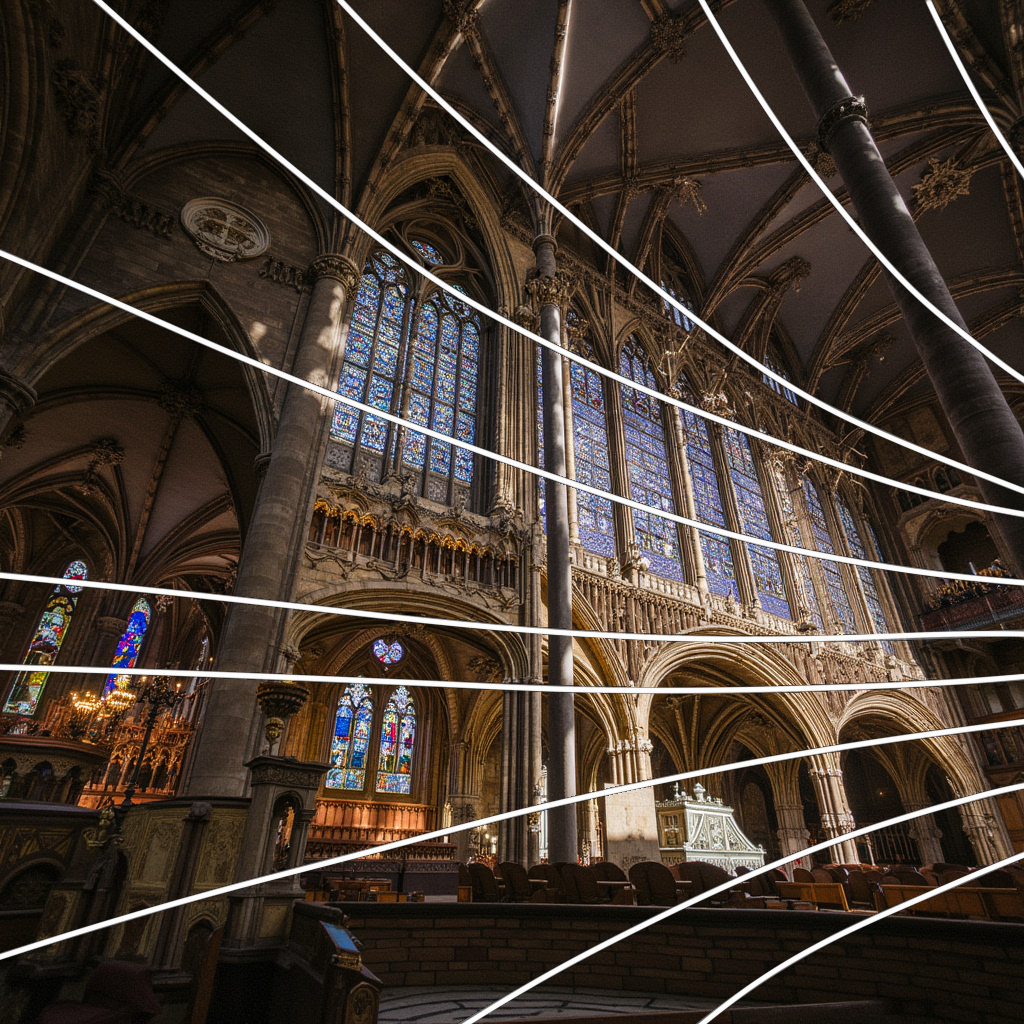}
        \caption{The Perspective ControlNet uses vanishing lines to define the 3D geometry and viewpoint of the scene.}
        \label{fig:teaser_perspective}
    \end{subfigure}
    \caption{Overview of the proposed ControlNets for high-level artistic control.}
    \label{fig:teaser}
\end{figure}

\begin{abstract}
\noindent While modern text-to-image diffusion models generate high-fidelity images, they offer limited control over the spatial and geometric structure of the output. To address this, we introduce and evaluate two ControlNets specialized for artistic control: (1) a proportion ControlNet that uses bounding boxes to dictate the position and scale of objects, and (2) a perspective ControlNet that employs vanishing lines to control the 3D geometry of the scene. We support the training of these modules with data pipelines that leverage vision-language models for annotation and specialized algorithms for conditioning image synthesis. Our experiments demonstrate that both modules provide effective control but exhibit limitations with complex constraints. Both models are released on HuggingFace: \url{https://huggingface.co/obvious-research}.
\end{abstract}

\section{Introduction}

Text-to-image diffusion models~\cite{ho2020denoising} have demonstrated remarkable capabilities in synthesizing complex images. However, their reliance on a single text prompt provides only coarse, global control, hindering their use in applications requiring compositional accuracy. The ControlNet architecture~\cite{zhang2023controlnet} introduced a paradigm for adding fine-grained, spatial conditioning to pre-trained models. While existing ControlNets for tasks like edge detection (Canny), line art, and depth provide low-level structural control, they don't provide control for all applications.

Our work focuses on providing artists with control based on fundamental principles of proportion and perspective. We extend this paradigm to a Flow Matching-based diffusion model, FLUX.1-dev~\cite{black2024flux}, and introduce two ControlNets, which are available open-source: the \href{https://huggingface.co/obvious-research/FLUX.1-dev-ControlNet-Proportion}{Proportion ControlNet} and the \href{https://huggingface.co/obvious-research/FLUX.1-dev-ControlNet-Perspective}{Perspective ControlNet}. A contribution is the development of data pipelines that generate the required training triplets from unannotated datasets. We evaluate each ControlNet individually and explore the dynamics of applying them simultaneously.

\section{Method}

For each control task, a dedicated ControlNet is trained while the pre-trained FLUX model remains frozen. The modules are initialized from a pre-trained \textit{LineArt} ControlNet to provide a foundational understanding of lines as structural guides. Training minimizes the conditional Flow Matching loss~\cite{lipman2023flow}, where the velocity field is conditioned on both text $c_{\text{text}}$ and a visual condition $c_{\text{cond}}$:
\[ \mathcal{L} = \mathbb{E}_{x_0, t, x_t, c_{\text{text}}, c_{\text{cond}}} \left[ \| v_t(x_t \mid x_0) - v_\theta(x_t, t, c_{\text{text}}, c_{\text{cond}}) \|^2 \right] \]

\subsection{High-Level Conditioning Design}
Our choice of conditioning inputs is motivated by the desire for artist-friendly, high-level control.

\paragraph{Proportion Control via Bounding Boxes.}
Our goal is to control the abstract proportions of a scene—the placement and visual weight of its semantic elements.
\begin{itemize}
    \item \textbf{Comparison to Regional Prompting:} Unlike regional prompting, which rigidly assigns different textual prompts to specific masked areas, our method uses a single global prompt. The bounding boxes define regions where elements described in the global prompt should appear, giving the model freedom to interpret the scene and decide how to fill those spaces. This decouples proportion from content segmentation.
    \item \textbf{Comparison to Low-Level Controls:} This approach also differs from pixel-level controllers like Canny or LineArt, which dictate the exact contours of an object. Bounding boxes are a higher level of abstraction, defining the semantic space an object should occupy, not its specific shape. This allows artists to easily apply proportion rules like the rule of thirds without pre-visualizing detailed outlines.
\end{itemize}

\paragraph{Perspective Control via Vanishing Lines.}
Our aim is to provide an intuitive method for defining the 3D geometry of a scene.
\begin{itemize}
    \item \textbf{Problem with Vanishing Points (VPs):} While VPs mathematically define perspective, they are a problematic conditioning input. VPs for parallel lines (1-point perspective) are at infinity and cannot be represented in a fixed-size image. Other VPs can be located far outside the canvas, making their encoding imprecise and difficult for a user to manipulate.
    \item \textbf{Solution with Vanishing Lines:} We therefore condition on vanishing lines. These lines are intuitive for an artist to draw, mimicking the sketching process of defining convergence. Crucially, they are always contained within the canvas, providing a direct, unambiguous, and spatially grounded proxy for VPs. This makes them a more robust and user-friendly input for controlling perspective.
\end{itemize}

\subsection{Data Pipelines}
\begin{figure}[t!]
    \centering
    \begin{subfigure}[b]{0.48\textwidth}
        \centering
        \begin{tikzpicture}[node distance=0.5cm]
        \node (wikiart) [data] {WikiArt Dataset};
        \node (filter) [process, below=of wikiart] {1. Aesthetic Filtering ($>$5.0)};
        \node (caption) [process, below=of filter] {2. Captioning (Florence-2)};
        \node (ground) [process, below=of caption] {3. Object Detection (Grounding DINO)};
        \node (triplet) [data, below=of ground] {Training Triplets};
        \node (train) [startstop, below=of triplet] {\textbf{FLUX ControlNet Proportion Training}};
        \draw [arrow] (wikiart) -- (filter); \draw [arrow] (filter) -- (caption);
        \draw [arrow] (caption) -- (ground); \draw [arrow] (ground) -- (triplet);
        \draw [arrow] (triplet) -- (train);
        \end{tikzpicture}
        \caption{Proportion pipeline.}
        \label{fig:compo_pipeline}
    \end{subfigure}
    \hfill
    \begin{subfigure}[b]{0.48\textwidth}
        \centering
        \begin{tikzpicture}[node distance=1cm]
        % Nodes
        \node (openimage) [data] {OpenImages v7 (1M subset)};
        \node (filter) [process, below of=openimage, yshift=-0.5cm] {1. Aesthetic Filtering ($>$3.5)};
        \node (vp_detect) [process, below of=filter, yshift=-0.5cm] {2. Vanishing Point Detection};
        \node (caption) [process, below of=vp_detect, yshift=-0.5cm] {3. Captioning (Florence-2)};
        \node (triplet) [data, below of=caption, yshift=-0.5cm] {Training Triplets};
        \node (train) [startstop, below of=triplet, yshift=-0.5cm] {\textbf{FLUX ControlNet Perspective Training}};
        
        % Arrows
        \draw [arrow] (openimage) -- (filter);
        \draw [arrow] (filter) -- (vp_detect);
        \draw [arrow] (vp_detect) -- (caption);
        \draw [arrow] (caption) -- (triplet);
        \draw [arrow] (triplet) -- (train);
    \end{tikzpicture}
    \caption{Perspective pipeline.}
\label{fig:persp_pipeline}
    \end{subfigure}
    \caption{Overview of the data pipelines. Both produce structured training triplets from unannotated datasets.}
    \label{fig:pipelines}
\end{figure}
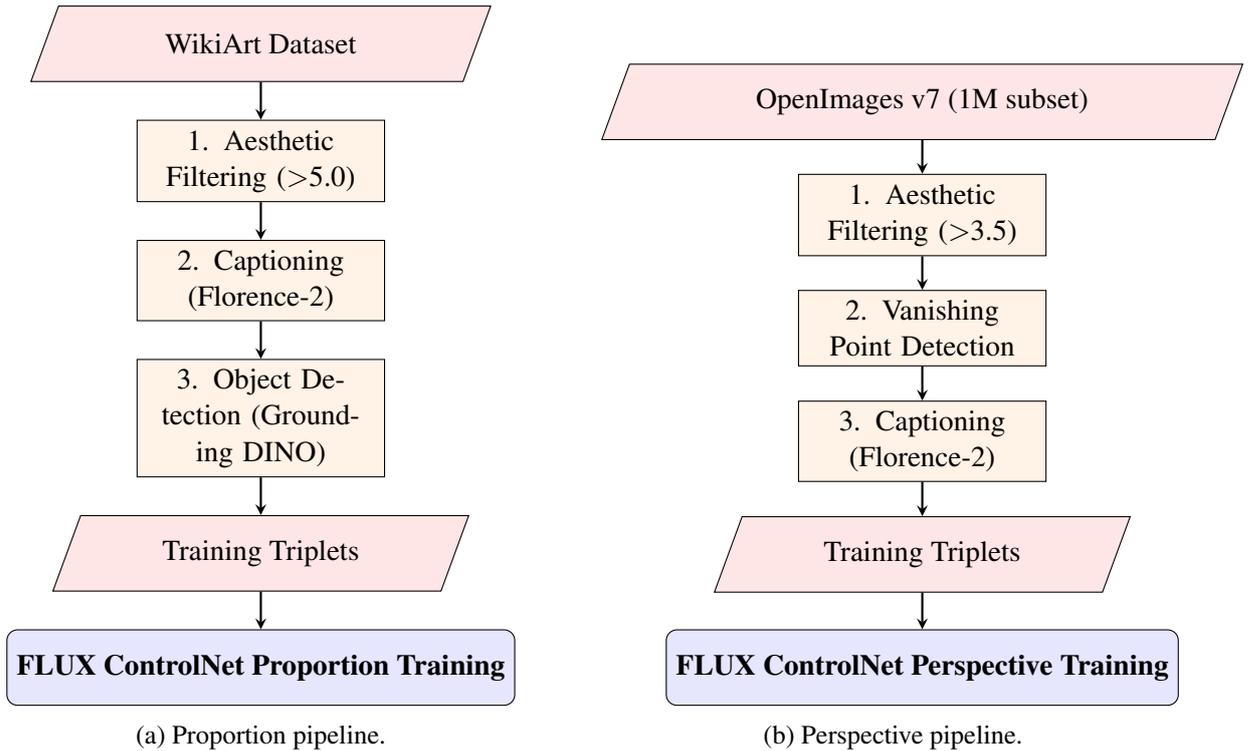

To train our models, we designed two distinct, fully automated data pipelines, shown in Figure \ref{fig:pipelines}.

\paragraph{Proportion Pipeline.} This pipeline processes the WikiArt dataset ($\sim$80k images). After filtering for aesthetic quality (score > 5.0) using a SigLIP-based~\cite{zhai2023siglip} predictor~\cite{discus0434aesthetic} to 73k images, Florence-2~\cite{xiao2023florence} generates a short caption (for the prompt) and a detailed caption. Grounding DINO~\cite{liu2023grounding} then uses the detailed caption to detect object bounding boxes, which are rendered to create the conditioning image.

\paragraph{Perspective Pipeline.} This pipeline processes a 1M image subset of OpenImages v7. After aesthetic filtering (score > 3.5), a geometric filter using the \textit{2-Line Exhaustive Search} algorithm~\cite{lu2017two} identifies images with a strong perspective structure. This yielded a final dataset of 373,632 images. The data distribution is heavily skewed: the vast majority have a reliable 1-point perspective, a smaller fraction have a 2-point perspective, and 3-point perspectives are extremely rare.

\section{Experiments and Results}

\begin{figure}[h!]
    \centering
    \begin{tabular}{ >{\centering\arraybackslash}p{0.55\textwidth} >{\centering\arraybackslash}p{0.35\textwidth} }
        \includegraphics[width=\linewidth]{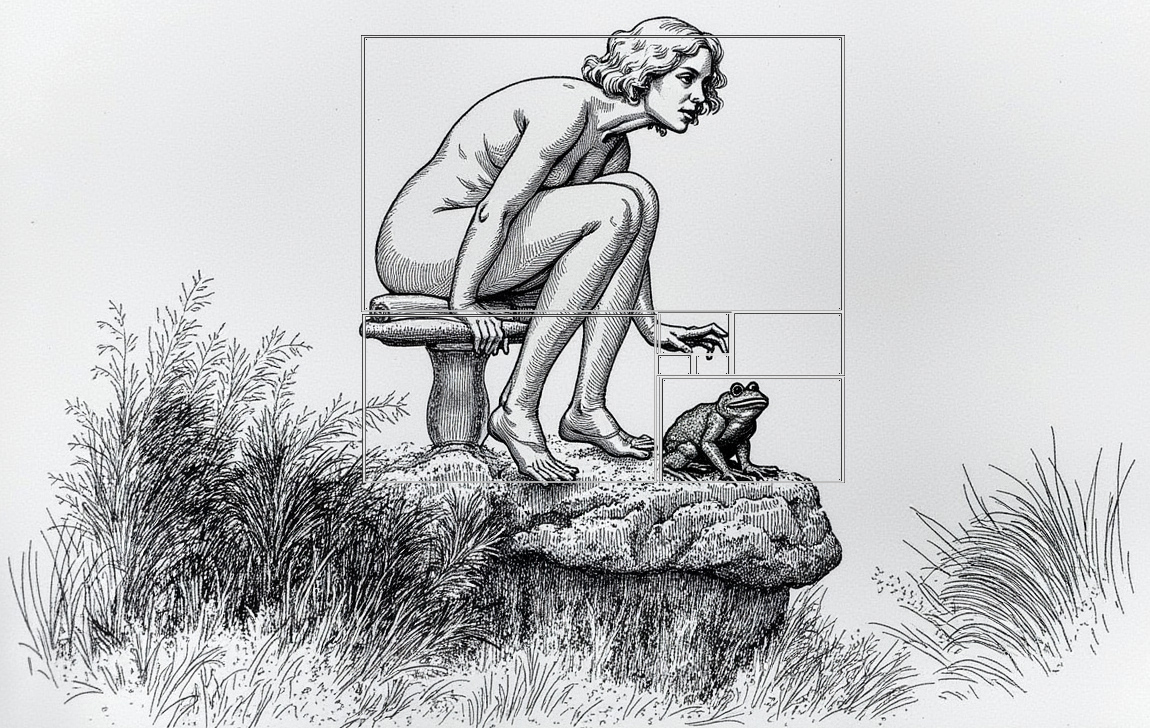} &
        \includegraphics[width=\linewidth]{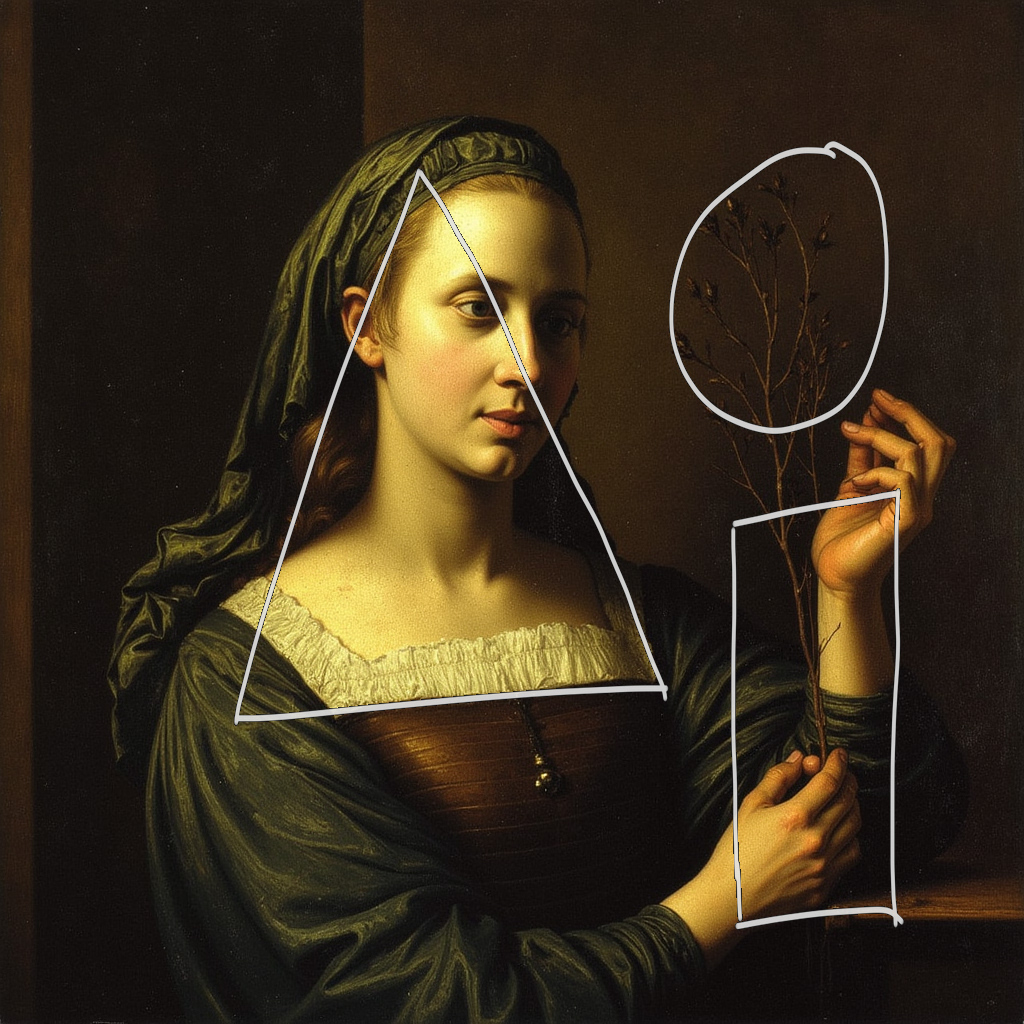} \\
        \small{(a) Bounding box control for object placement.} & 
        \small{(b) Emergent generalization to non-rectangular shapes.} \\
    \end{tabular}
    \caption{Results from the Proportion ControlNet.}
    \label{fig:proportion_results}
\end{figure}

\subsection{Qualitative Results of Individual Modules}

\begin{figure}[h!]
    \centering
    \includegraphics[width=0.4\textwidth]{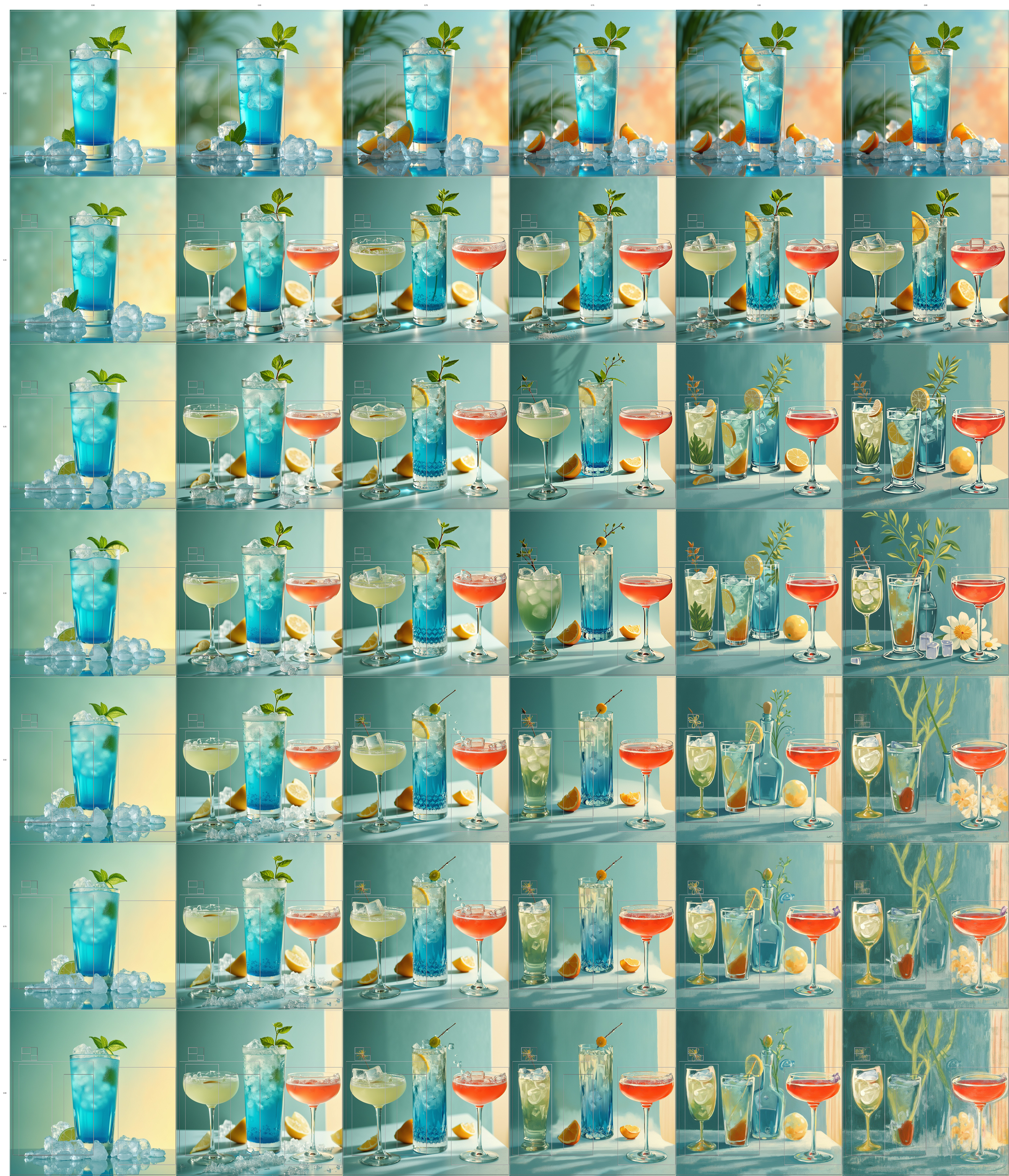}
    \caption{Illustration of the stylistic bias. The horizontal axis represents the ControlNet guidance strength (0.5 to 0.9). The vertical axis represents the end percentage of the diffusion process during which the ControlNet is active (0.1 to 0.9). The "pictorial" effect intensifies as both values increase.}
    \label{fig:style_grid}
\end{figure}

\paragraph{Proportion Control.} The model effectively adheres to bounding box constraints (Figure \ref{fig:proportion_results}a). Training on WikiArt induced a strong "pictorial" style bias, whose intensity correlates with higher ControlNet guidance strength (Figure \ref{fig:style_grid}). Due to its LineArt initialization, the model exhibits an emergent ability to interpret non-rectangular shapes as proportional guides (Figure \ref{fig:proportion_results}b).

\begin{figure}[ht!]
    \centering
    \begin{tabular}{ >{\centering\arraybackslash}p{0.35\textwidth} >{\centering\arraybackslash}p{0.35\textwidth} }
        \includegraphics[width=\linewidth]{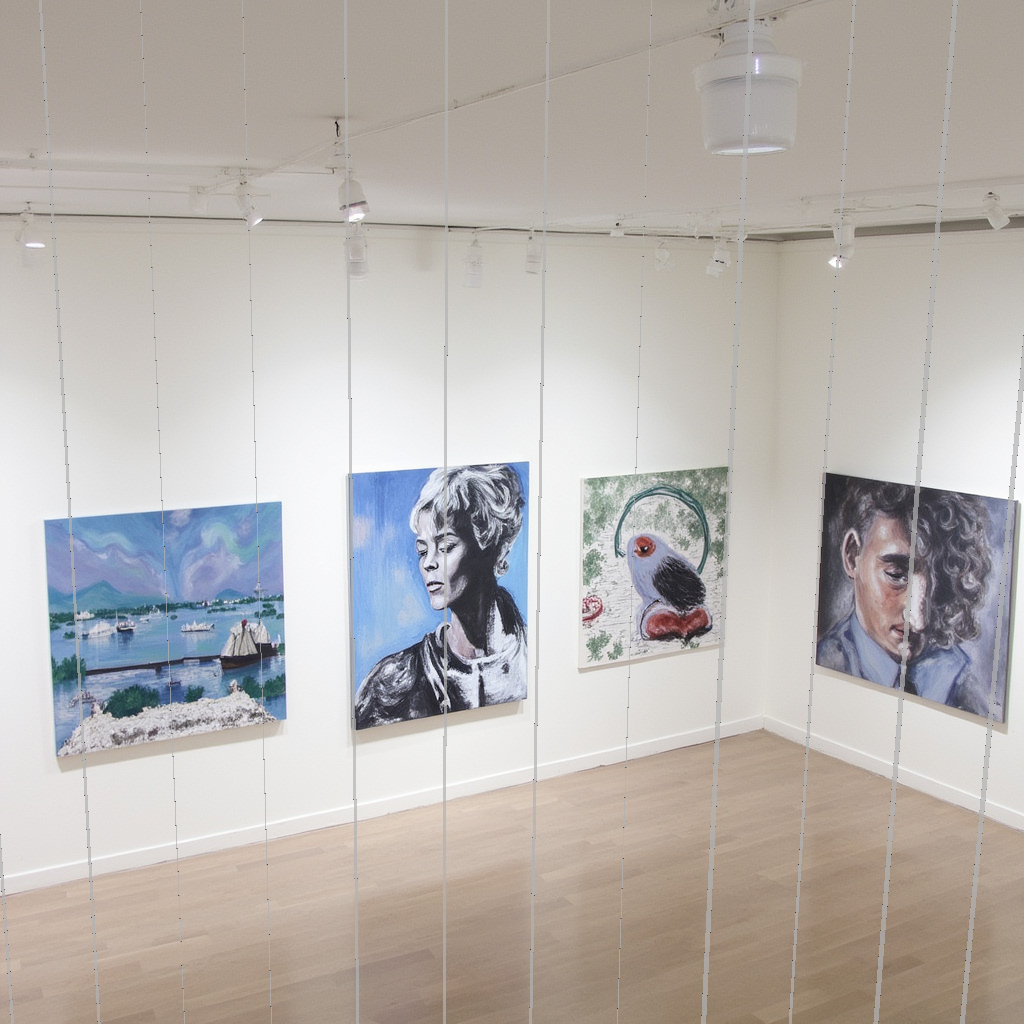} &
        \includegraphics[width=\linewidth]{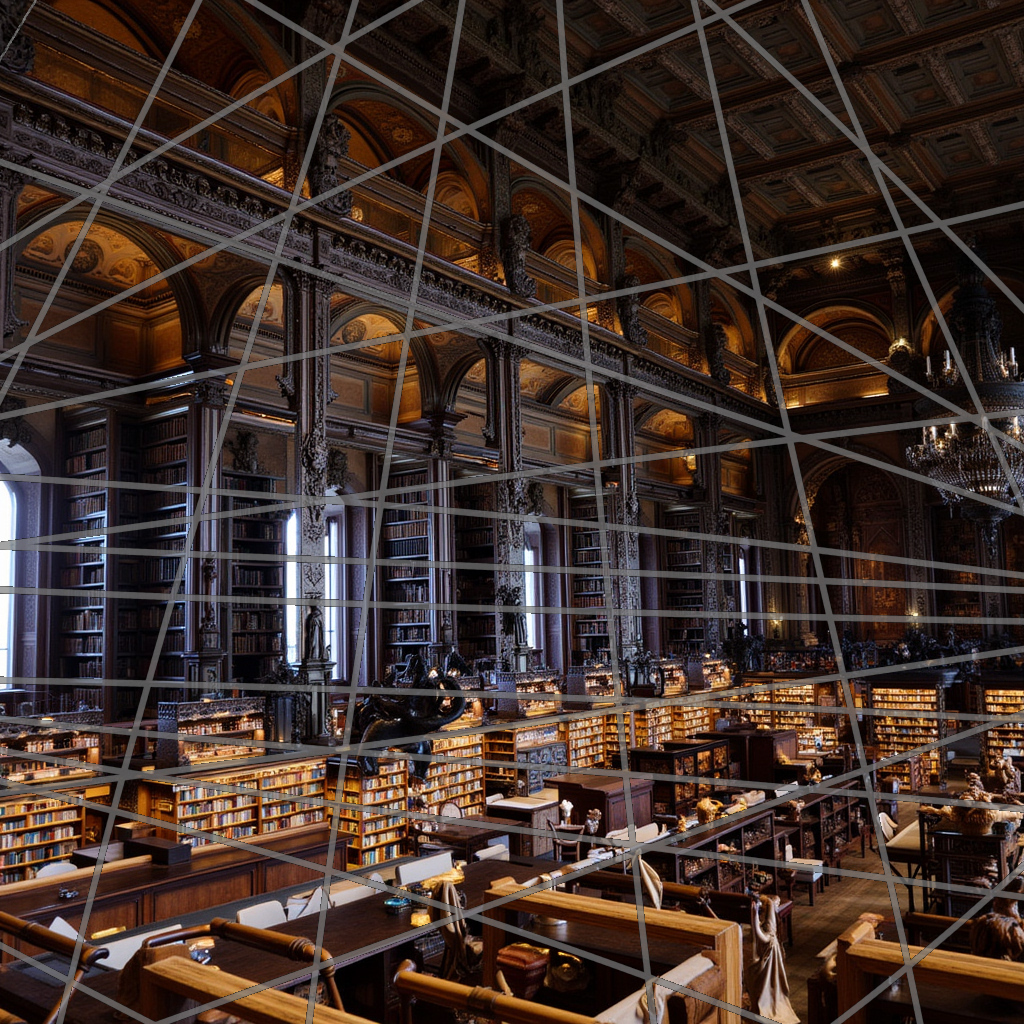} \\
        \small{(a) Successful control for a 1-point perspective.} & 
        \small{(b) Failure on a 3-point perspective ("axis dropping").} \\
    \end{tabular}
    \caption{Results from the perspective ControlNet.}
    \label{fig:persp_results}
\end{figure}

\begin{figure}[h!]
    \centering
    \begin{tabular}{ >{\centering\arraybackslash}p{0.35\textwidth} >{\centering\arraybackslash}p{0.35\textwidth} }
        \includegraphics[width=\linewidth]{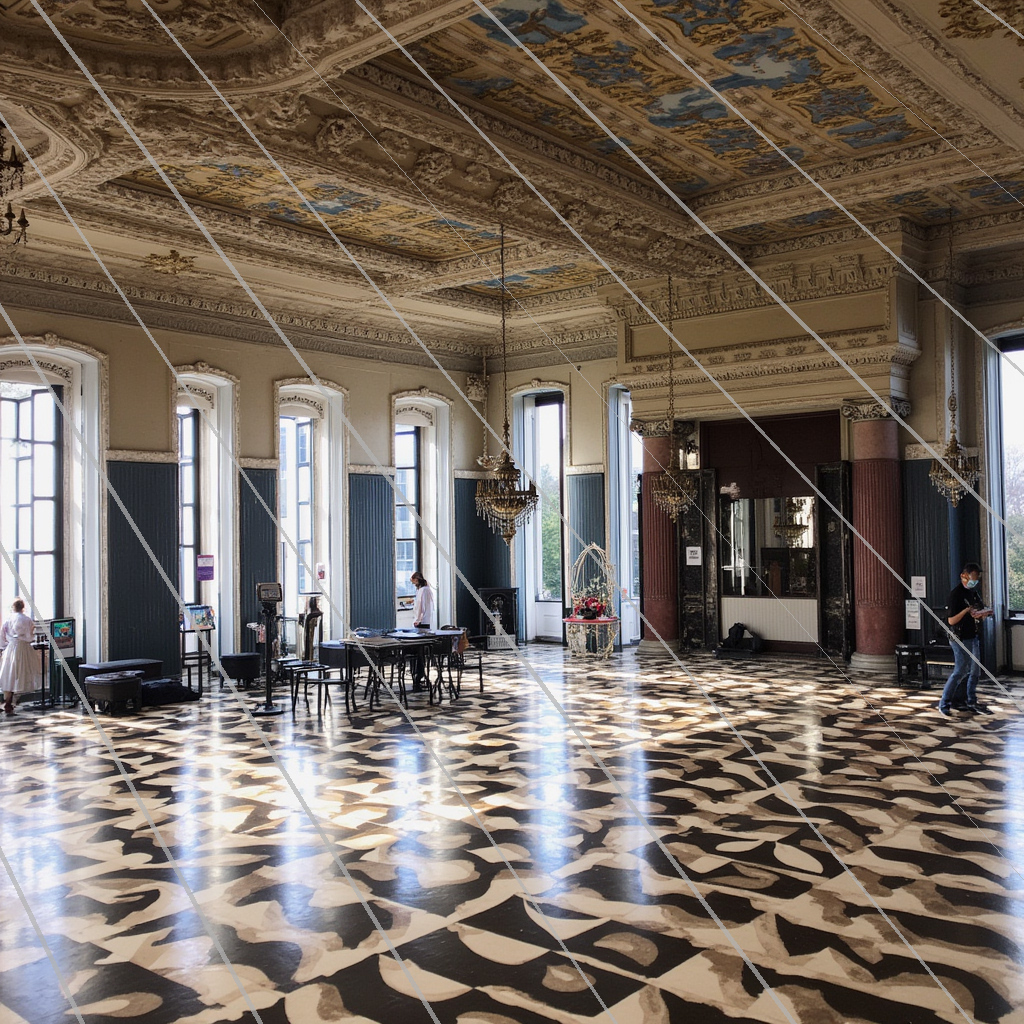} &
        \includegraphics[width=\linewidth]{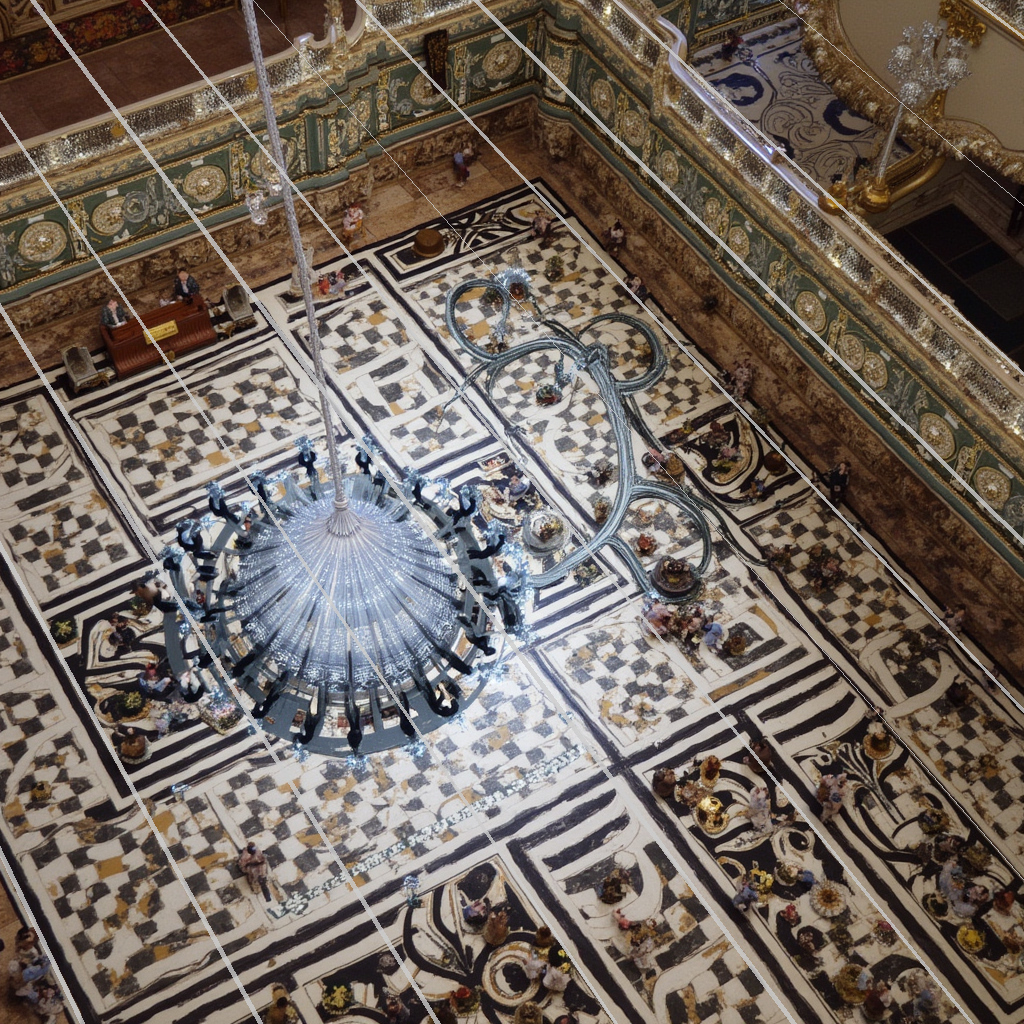} \\
        \small{(a) Default prompt ignores tilted perspective.} & 
        \small{(b) Explicit "Top view" prompt forces adherence.} \\
    \end{tabular}
    \caption{Overcoming model priors for non-standard perspectives requires textual guidance.}
    \label{fig:dutch_angle}
\end{figure}

\paragraph{Perspective Control.} The model successfully generates scenes respecting 1- and 2-point perspectives (Figure \ref{fig:persp_results}a). However, it consistently fails to render 3-point perspectives, often exhibiting "axis dropping" by ignoring vertical convergence (Figure \ref{fig:persp_results}b). This failure is directly attributable to the skewed data distribution. Furthermore, the model shows a strong prior for straight horizons, likely inherited from both its training data and the base FLUX model. Overcoming this for non-standard views (e.g., Dutch angles) requires explicit textual prompting (Figure \ref{fig:dutch_angle}).

\subsection{Multi-ControlNet Composition}
\begin{figure}[h!]
    \centering
    \begin{tabular}{ >{\centering\arraybackslash}p{0.35\textwidth} >{\centering\arraybackslash}p{0.35\textwidth} }
        \includegraphics[width=\linewidth]{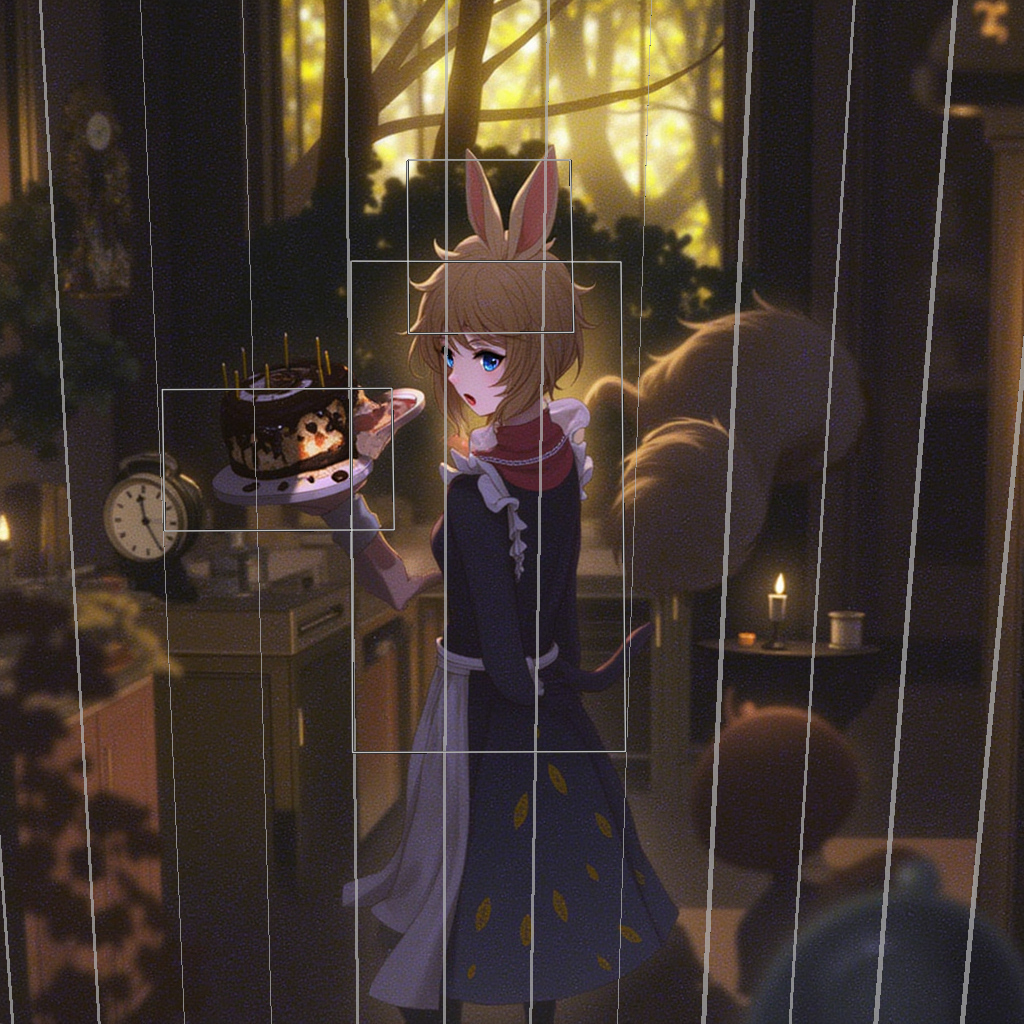} &
        \includegraphics[width=\linewidth]{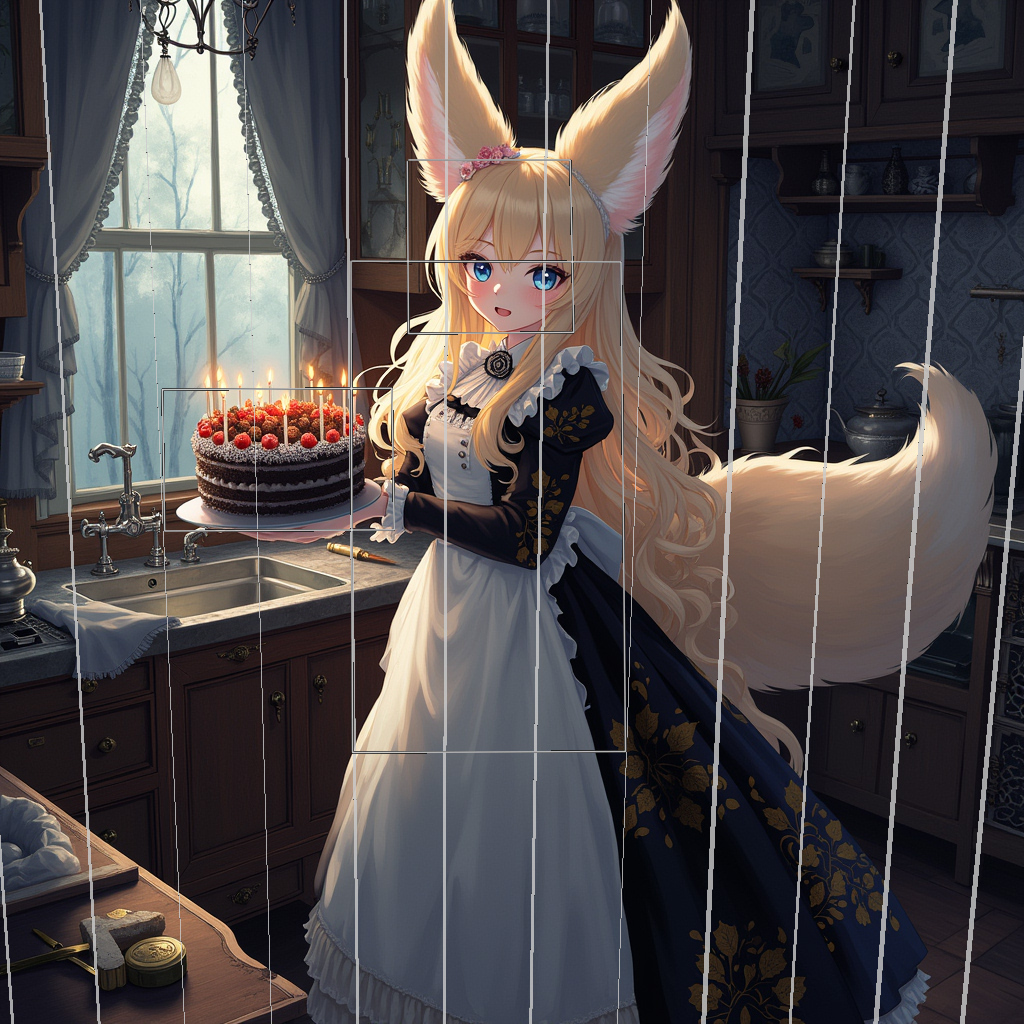} \\
        \small{(a) High guidance (0.8 each) leads to severe color artifacts and loss of detail.} & 
        \small{(b) A scene with partial adherence to both controls with low guidance (0.5 each).} \\
    \end{tabular}
    \caption{Results from multi-ControlNet composition.}
    \label{fig:multi_control}
\end{figure}

We investigated applying both ControlNets simultaneously. Using guidance strengths optimal for a single module (e.g., 0.8) leads to image degradation. While structurally recognizable, the outputs suffer from severe color artifacts and "mushy" textures where details become incoherent (Figure \ref{fig:multi_control}a). Stable generation requires reducing the guidance strength of each module to approximately 0.5. At this level, it is possible to obtain results that partially adhere to both constraints (Figure \ref{fig:multi_control}b). However, robust and precise combined control is difficult to achieve consistently.

\section{Conclusion}
We have presented two specialized ControlNet modules for the FLUX architecture that provide effective proportion and perspective control. Future work should focus on improving data diversity through synthetic generation (e.g., from 3D scenes to create balanced perspective datasets) and data augmentation. 

\newpage
\bibliographystyle{plain}
\bibliography{references}

\newpage
\begin{appendices}
\begin{figure}[h]
    \centering
    \includegraphics[width=1\linewidth]{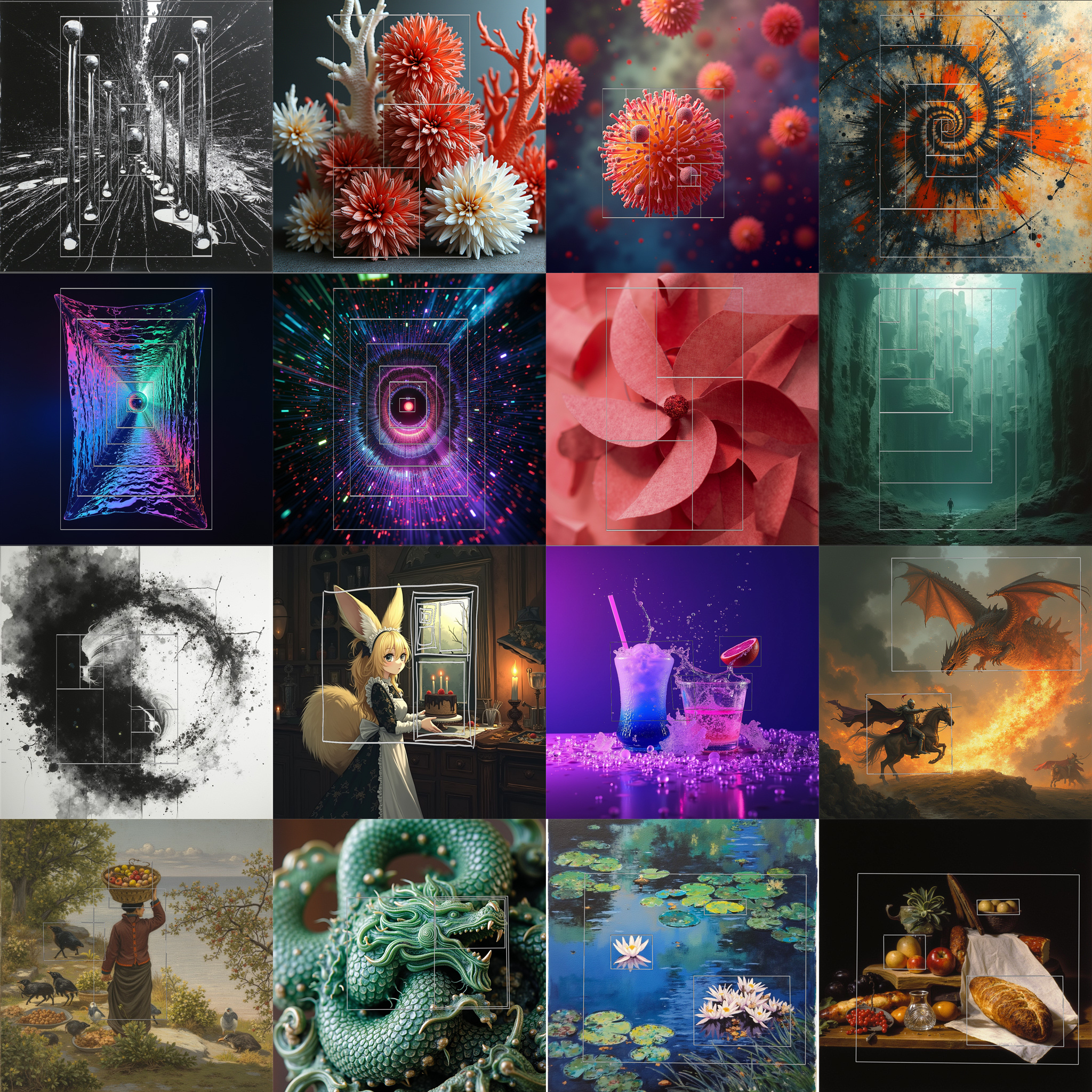}
    \caption{Examples of generations with the Proportion ControlNet. The conditioning is overlayed on the generated image.}
\end{figure}

\begin{figure}
    \centering
    \includegraphics[width=1\linewidth]{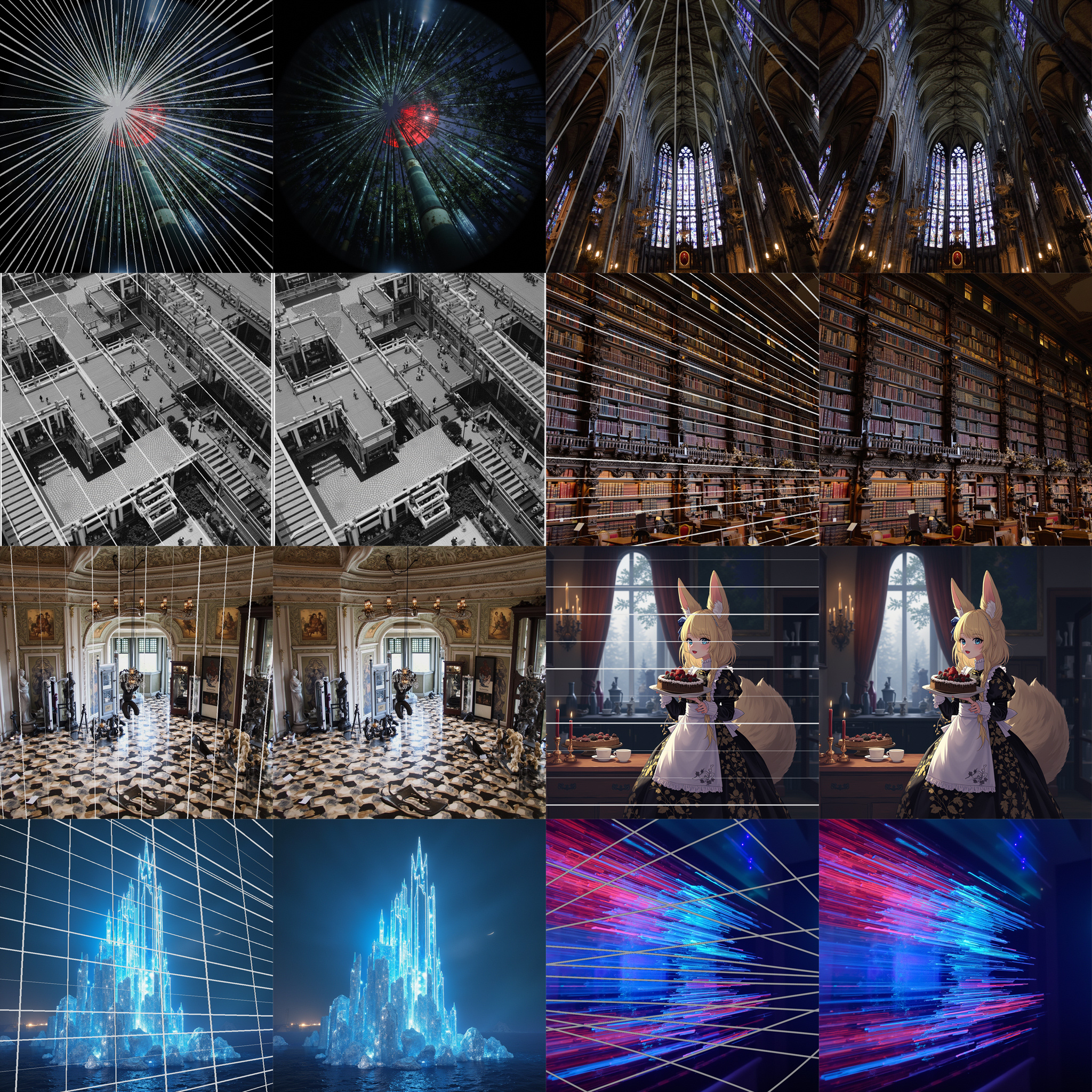}
    \caption{Examples of generations with the Perspective ControlNet. The conditioning is overlayed on the generated image on the left.}
\end{figure}
\end{appendices}

\end{document}